%% file: main.tex
\documentclass[10pt,twocolumn,letterpaper]{article}

 \usepackage[pagenumbers]{cvpr} 
\usepackage{float} 

\usepackage[accsupp]{axessibility}


\definecolor{cvprblue}{rgb}{0.21,0.49,0.74}
\usepackage[pagebackref,breaklinks,colorlinks,allcolors=cvprblue]{hyperref}
\usepackage{algorithm}

\usepackage{algorithmic}

\usepackage{booktabs}
\usepackage{multirow}
\usepackage{graphicx}
\usepackage{colortbl}
\def\paperID{4699} 
\def\confName{CVPR}
\def\confYear{2025}

\title{V2X-R: Cooperative LiDAR-4D Radar Fusion \\  with Denoising Diffusion for 3D Object Detection}

\author{Xun Huang\textsuperscript{\rm 1}\textsuperscript{\rm 2}\textsuperscript{\rm 3} \quad Jinlong Wang\textsuperscript{\rm 1}\textsuperscript{\rm 2} \quad Qiming Xia\textsuperscript{\rm 1}\textsuperscript{\rm 2} \quad   Siheng Chen\textsuperscript{\rm 4} \\ Bisheng Yang\textsuperscript{\rm 5} \quad Xin Li\textsuperscript{\rm 6} \quad Cheng Wang\textsuperscript{\rm 1}\textsuperscript{\rm 2} \quad Chenglu Wen\textsuperscript{\rm 1}\textsuperscript{\rm 2}\thanks{Corresponding author, {\tt\small clwen@xmu.edu.cn}} \\
    \textsuperscript{\rm 1}Fujian Key Laboratory of Sensing and Computing for Smart Cities, Xiamen University, China \\
    \textsuperscript{\rm 2}Key Laboratory of Multimedia Trusted Perception and Efficient Computing, \\Ministry of Education of China, Xiamen University, China \\
    \textsuperscript{\rm 3}Zhongguancun Academy
    \textsuperscript{\rm 4}Shanghai Jiao Tong University \\
    \textsuperscript{\rm 5}Wuhan University 
    \textsuperscript{\rm 6}Texas A\&M University
\\
}

\begin{document}
\maketitle
\input{sec/0_abstract}    
\input{sec/1_intro}
\input{sec/2_related_work}
\input{sec/3_V2Xdataset}

\input{sec/4_Method}

\input{sec/5_Experiments}
\input{sec/6_Conclusion}
\input{sec/X_suppl}

{
    \small
    \bibliographystyle{ieeenat_fullname}
    \bibliography{main}
}


\end{document}

%% file: sec/0_abstract.tex
\begin{abstract}
Current Vehicle-to-Everything (V2X) systems have significantly enhanced 3D object detection using LiDAR and camera data. However, they face performance degradation in adverse weather. 
Weather-robust 4D radar, with Doppler velocity and additional geometric information, offers a promising solution to this challenge. 
To this end, we present V2X-R, the first simulated V2X dataset incorporating LiDAR, camera, and 4D radar modalities. 
V2X-R contains 12,079 scenarios with 37,727 frames of LiDAR and 4D radar point clouds, 150,908 images, and 170,859 annotated 3D vehicle bounding boxes. Subsequently, we propose a novel cooperative LiDAR-4D radar fusion pipeline for 3D object detection and implement it with multiple fusion strategies. 
To achieve weather-robust detection, we additionally propose a Multi-modal Denoising Diffusion (MDD) module in our fusion pipeline. MDD utilizes weather-robust 4D radar feature as a condition to guide the diffusion model in denoising noisy LiDAR features.
Experiments show that our LiDAR-4D radar fusion pipeline demonstrates superior performance in the V2X-R dataset. Over and above this, our MDD module further improved the foggy/snowy performance of the basic fusion model by up to  5.73\%/6.70\% and barely disrupting normal performance. The dataset and code will be publicly available at: \href{https://github.com/ylwhxht/V2X-R}{https://github.com/ylwhxht/V2X-R}.
\end{abstract}

%% file: sec/1_intro.tex
\section{Introduction}
\label{sec:intro}
Autonomous driving and other unmanned systems have garnered widespread attention in recent years. This has led to rapid advancements in 3D object detection~\citep{bevfusion,safdnet,cmd}.
Outdoor environments, however, present complex and dynamic challenges, including various occlusions and weather conditions \cite{cpd,PTT}. Such factors significantly impact the performance of 3D object detection. Consequently, some effort has been made to explore multi-agent cooperative perception, such as vehicle-to-vehicle (V2V), vehicle-to-infrastructure (V2I), and vehicle-to-everything (V2X)  \citep{opv2v,v2xvit,xu2023v2v4real,DAIR-V2X}. Benefiting from the information shared between agents, in complex outdoor scenarios, cooperative 3D object detection has natural advantages, such as long detection distance and multi-view object observation. 

\begin{figure}[!t]
  \includegraphics[width=1.0\columnwidth]{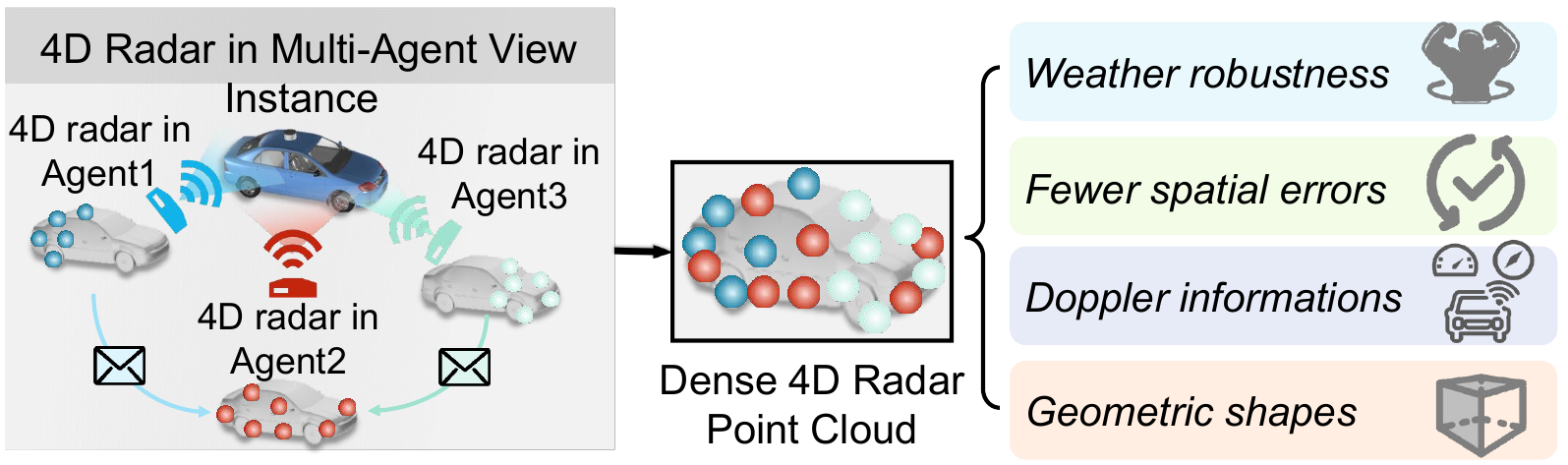}
  \caption{The advantages of the dense 4D radar point cloud in multi-agent view. Including weather robustness, fewer spatial errors, Doppler information, and geometric shapes.}
  \label{intro}
\end{figure}

Current research in cooperative 3D object detection mainly focuses on two strategies: LiDAR-based single modality \citep{cobevt,v2xvit,where2comm, scope, sicp} and LiDAR-camera multi-modal fusion \citep{opv2v,hmvit,bm2cp,codefilling,coopdet3D}. 
The latter strategy provides more fine-grained information and, therefore, improves the performance of single LiDAR-based methods to some extent.
However, both LiDAR point clouds and camera images are weather-sensitive. They are all prone to become noisy in adverse weather \citep{fogsim,SRKD,snowsim}.

Aside from LiDAR and camera, 4D radar has also been widely noticed \cite{L4DR,ourl4dr,survey_4d1,survey_4d2,InterFusion,MMFusion} because it can perform all-weather sensing and provide speed measurements \cite{RCFusion}. Fusing 4D radar and LiDAR is expected to improve the perception performance under adverse weather conditions.
As shown in Fig. \ref{intro}, cooperative 4D radar and LiDAR fusion perception have the following advantages:
\begin{itemize}
\item \textit{Weather robustness.} The millimeter-wave signals of 4D radar easily penetrates particles in adverse weather \cite{MVDNet,robots1,robots2}. Fusing 4D radar and LiDAR will grant the model weather robust sensing capabilities.
\item \textit{Fewer spatial relationship errors.} The error-prone operations (view-transformation or depth-estimation) \citep{virconv,ted,L4DR} are not involved in the process of 4D radar and LiDAR point cloud fusing. Thus, there are fewer corresponding spatial relationship errors in fusing 4D radar and LiDAR. 
\item \textit{Additional Doppler information.} The Doppler information provided by 4D radar is favorable for object detection \citep{ourl4dr,L4DR},  significantly aiding in the detection of moving objects.
\item \textit{Richer geometric shape information.} The multi-agent view can significantly address the limitation of low resolution in 4D radar. This empowers 4D radar's capacity to complement LiDAR with more geometric shape information of objects.
\end{itemize}



However, there is a lack of 4D radar data in the current cooperative perception dataset. Therefore, we present V2X-R, the first simulated cooperative 3D Object Detection V2X dataset that not only includes LiDAR, cameras, but also 4D radar data. V2X-R contains 12,079 scenarios with 37,727 frames of LiDAR and 4D radar point clouds, 150,908 images, and 170,859 annotated 3D vehicle bounding boxes. 
Built upon this dataset, we develop a general cooperative LiDAR-4D radar fusion pipeline for 3D object detection. The entire pipeline consists of four stages: 1) Encode by each agent. 2) Agent fusion. 3) Modal fusion. 4) Box prediction. To address the challenge of agent-fused LiDAR features becoming noisy in adverse weather, we propose a novel Multi-modal Diffusion Denoising (MDD) module in the modal fusion stage of the pipeline. MDD transforms the noise feature distribution into the easy-to-fit Gaussian distribution by reparameterization, which solves the challenge of complex and variable weather noise features that are difficult to fit. It further utilizes weather-robust 4D radar feature as a condition to prompt the diffusion model to denoise noisy LiDAR features. Notably, our MDD module barely disrupts performance in normal weather, attaining high performance in both normal and adverse weather. 

\begin{figure}[!t]
  \includegraphics[width=1.0\columnwidth]{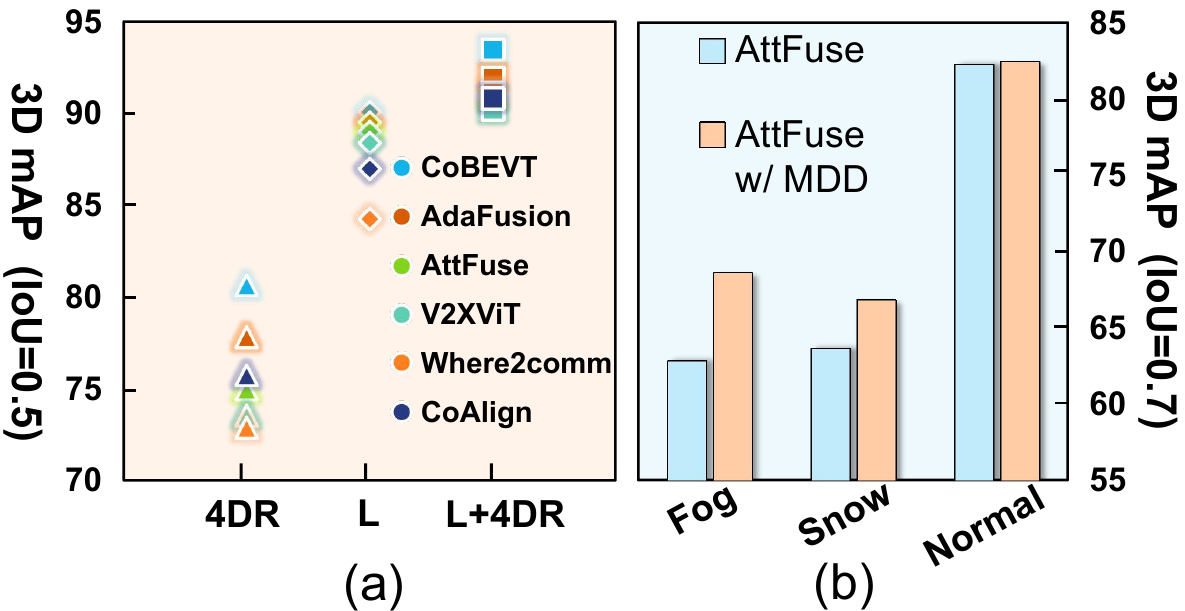}
  \caption{The performance of different methods in our V2X-R dataset. (a) Performance comparison of different modalities (L and 4DR represent LiDAR and 4D radar modality, respectively). (b) Performance comparison of Attfuse \citep{opv2v} model (with and without our MDD module) under different weather conditions.}
  \label{intro_performance}
\end{figure}

 Subsequently, we implement the cooperative LiDAR-4D radar fusion pipeline with various agent fusion and modal fusion strategies on our V2X-R dataset, establishing a comprehensive benchmark for cooperative 3D object detection. 
 The comprehensive experiment results show that the LiDAR-4D radar fusion demonstrates superior performance based on various model architectures, as shown in Fig.~\ref{intro_performance}(a). The effectiveness of our designed MDD module is also validated under noise-prone foggy and snowy weather. As shown in Fig.~\ref{intro_performance}(b), by incorporating our MDD module, AttFuse \citep{opv2v} has significantly improved the weather-robustness ability. 
 
 Our contributions can be summarized in three key points:
 
\begin{itemize}
\item We present V2X-R, the first simulated V2X dataset that not only includes LiDAR, cameras, but also 4D radar data. This dataset lays the data foundation for research in V2X cooperative perception with 4D radar.

\item We designed a novel Multi-modal Diffusion Denoising (MDD) module to utilize reparameterization and weather-robust 4D radar feature to handle hard-to-fit noisy LiDAR feature. Our MDD shows effectiveness in both single and multi-agent, simulated and real scenarios.

\item We construct a cooperative LiDAR-4D radar fusion pipeline for 3D object detection. We have implemented this pipeline with various fusion strategies and provided a comprehensive benchmark on our V2X-R dataset, boosting the research in cooperative LiDAR-4D radar fusion for cooperative 3D object detection.
\end{itemize}

%% file: sec/2_related_work.tex
\section{Related Work}

\subsection{3D Object Detection}
3D object detection has garnered significant attention, with mainstream approaches categorized as LiDAR-based \citep{second, PointPillars, 3DSSD, dsvt, pvrcnn, pvrcnn++, coin, voxelrcnn, safdnet, hinted} or LiDAR-camera fusion-based \citep{virconv, ted, bevfusion, cmt, DeepFusion}. The latter achieves superior performance by combining LiDAR's depth and geometry information with the fine-grained details from images \cite{lin2024boosting, lin2024ssmae, lin2024sparse}. However, LiDAR's weather sensitivity leads to poor performance in adverse conditions, prompting research into weather-robust 3D object detection methods \citep{SRKD, snowod, fogsim}. Some approaches utilize weather-robust 3D Radar \citep{MVDNet, LiRaFusion, ST-MVDNet, ST-MVDNet++} or 4D radar \citep{InterFusion, MMFusion, L4DR, ourl4dr} fusion with LiDAR. 4D radar has received more attention due to higher resolution and elevation, but cooperative LiDAR-4D radar fusion remains unexplored.

\subsection{Cooperative Perception Datasets}
Large-scale multi-sensor cooperative perception datasets, such as V2V4Real \citep{V2V4Real}, are crucial for training robust and generalizable perception models. However, real-world data collection is costly and time-consuming. Simulation platforms like CARLA \citep{carla}, SUMO \citep{SUMO}, and OpenCDA \citep{opencda} offer a cost-effective alternative. Datasets such as OPV2V \citep{opv2v} and V2X-Sim \citep{v2x-sim} have been developed using these tools, supporting tasks like 3D object detection, tracking, and bird's-eye view (BEV) segmentation. However, above datasets lack of 4D Radar data, which limits the research of model robustness under adverse weather conditions.

\subsection{V2X Cooperative Perception}
V2X cooperative perception enhances semantic understanding by sharing observations from surrounding vehicles and infrastructure. Research can be categorized into early fusion \citep{Cooper}, mid fusion \citep{opv2v, scope, cobevt, v2xvit, where2comm}, and late fusion \citep{Car2X, Rawashdeh_Wang_2018, DAIR-V2X}. Mid fusion, involving feature sharing among connected and automated vehicles (CAVs), is favored for its performance-bandwidth balance \citep{scope}. Most methods are LiDAR-based, with advancements such as self-attention for feature fusion \citep{opv2v}, unified Transformer structures for heterogeneous data \citep{v2xvit}, and bandwidth optimization \citep{where2comm}. For LiDAR-camera fusion, BM2CP \citep{bm2cp} proposes a multi-modal cooperative perception framework, while CodeFilling \citep{codefilling} facilitates perception-communication trade-offs. However, the weather sensitivity of LiDAR and camera sensors 
still poses challenges for multi-agent 3D object detection.

%% file: sec/3_V2Xdataset.tex
\section{V2X-R Dataset}
\label{sec:dataset}
\subsection{Simulator Selection}
We chose CARLA \cite{carla} as the primary simulator for data collection. Our data is derived from the eight towns provided by CARLA. However, since CARLA lacks vehicle-to-everything (V2X) communication and cooperative driving capabilities, we used OpenCDA \cite{opencda} integrated with CARLA, a cooperative simulation platform that supports multiple cooperative agents and basic control over embedded vehicular network communication protocols, to generate our V2X-R dataset. Additionally, we integrated the simulated 4D radar sensor into the OpenCDA framework for 4D radar simulation and data collection.

\subsection{Sensor configuration}
As described in Table \ref{tab:Sensor details}, each Connected Autonomous Vehicle (CAV) and Infrastructure is equipped with four cameras; a 64-channel LiDAR sensor featuring a detection range of 120 meters; a radar sensor with a vertical field of view of 30$^{\circ}$ and a maximum detection range of 150 meters. The vehicle is equipped with a Global Navigation Satellite System (GNSS) with an altitude noise of 0.001 meters, while the roadside unit (RSU) has an altitude noise of 0.05 meters. Finally, based on the above configuration, our V2X-R contains a total of 12,079 scenarios with 37,727 frames of LiDAR and 4D radar point clouds, 150,908 images, and 170,859 annotated 3D vehicle bounding boxes.

\begin{table}[!t]
\resizebox{\columnwidth}{!}{%
\begin{tabular}{c|c}
\bottomrule
\textbf{Sensors}                  & \textbf{Details}                                                                                                                                                                                  \\ \hline
{\color[HTML]{08090C} 4x Camera}  & {\color[HTML]{08090C} \begin{tabular}[c]{@{}c@{}}4 units RGB,Positions:\\(2.5,0,1.0,0),(0.0,0.3,1.8,100), \\ (0.0,-0.3,1.8,-100), (-2.0,0.0,1.5,180)\end{tabular}}                                  \\
\hline
{\color[HTML]{08090C} 1x LiDAR}   & {\color[HTML]{08090C} \begin{tabular}[c]{@{}c@{}}64 channels,120m range, -25$^{\circ}$to 2$^{\circ}$ vertical FOV,\\ 0.02 noise standard deviation, 20 Hz rotation frequency\end{tabular}} \\
\hline
{\color[HTML]{08090C} 1x 4D radar} & {150m range, 120$^{\circ}$ horizontal FOV, 30$^{\circ}$ vertical FOV}                                                                                     \\ \hline
{\color[HTML]{08090C} GPS \& IMU} & {\color[HTML]{08090C} \begin{tabular}[c]{@{}c@{}}Vehicle GNSS:altitude noise 0.001m; \\Vehicle IMU: heading noise 0.1$^{\circ}$,  speed noise 0.2m/s; \\RSU GNSS:altitude noise 0.05 m,\end{tabular}}          \\ \toprule
\end{tabular}%
}
\caption{Sensor configuration details of our V2X-R dataset.}
\label{tab:Sensor details}
\end{table}
\begin{figure}[!t]
  \includegraphics[width=1.0\columnwidth]{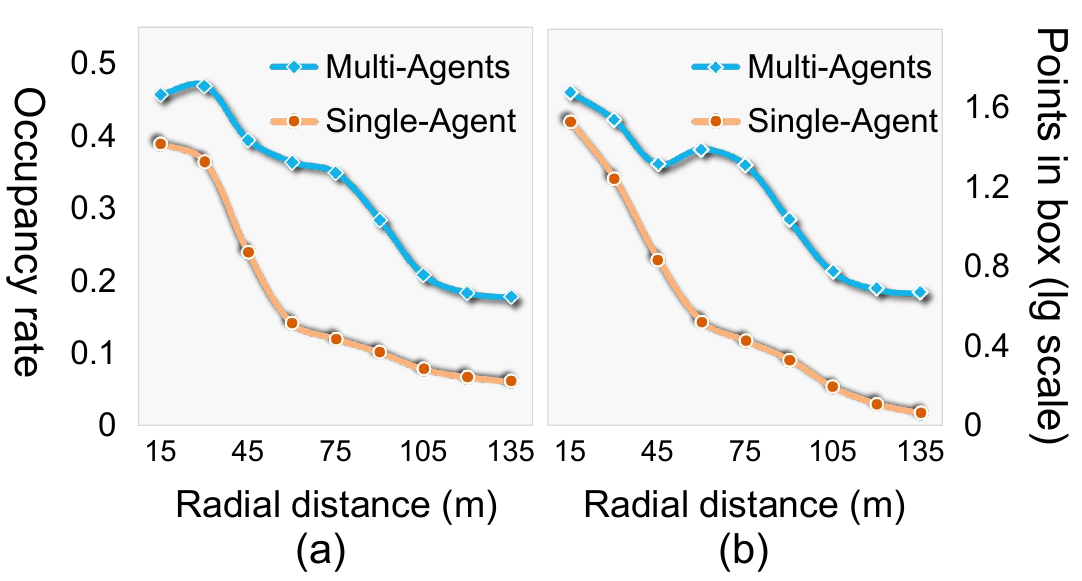}
  \caption{4D radar point cloud occupancy rate (a) and number of points (b) within the ground truth bounding boxes for radial distance from ego vehicles.}
  \label{radar_vis}
\end{figure}

\subsection{Cooperative 4D Radar Point Cloud Analysis}
We analyzed the instance occupancy and points of the 4D radar point cloud before and after the multi-agent cooperative communication, as shown in Fig. \ref{radar_vis}. The 4D radar point cloud instance occupancy and points at different distances increase significantly with multi-agent collaboration. Especially in the middle and long distance, 4D radar point clouds can still have relatively high instance occupancy and points. 
These data demonstrate the independent perception capability of the cooperative 4D radar point cloud.

\begin{figure*}[!t]
\centering
  \includegraphics[width=0.9\textwidth]{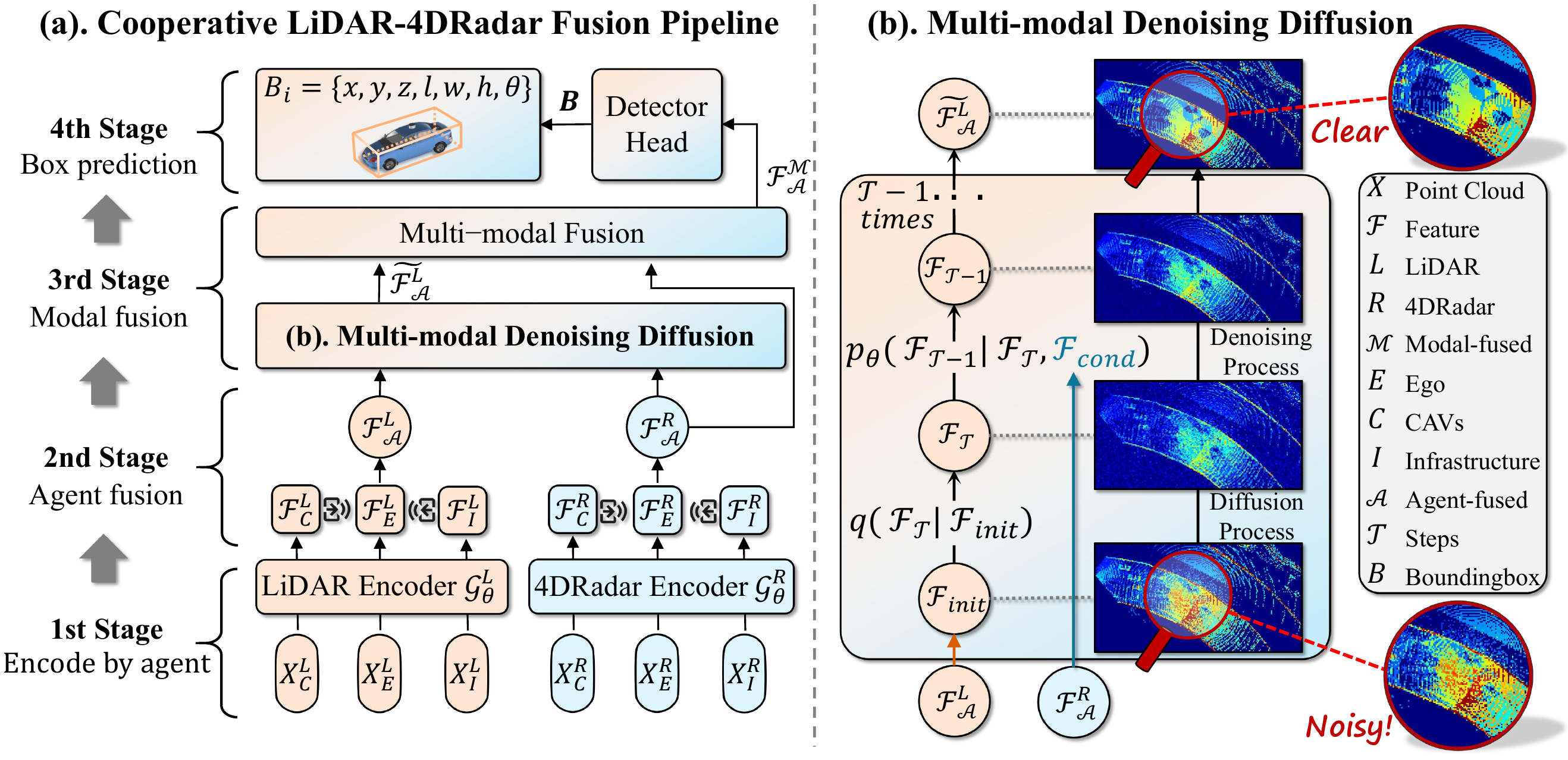}
  \caption{The pipeline of constructed cooperative LiDAR-4D radar fusion for weather-robust 3D object detection. The fusion pipeline (a) is first fed with multi-modal point cloud (LiDAR and 4D radar) from multi-agent: Ego Vehicle (E), Connected Automated Vehicle (C), and Infrastructure (I). Subsequent fusion consists of four stages: 1) Encode by each agent. 2) Agent fusion. 3)Modal fusion. 4) Box prediction. Additionally, a Multi-modal Denoising Diffusion (b) module is integrated into the pipeline to achieve weather-robust ability. The input noisy LiDAR features are first subjected to a diffusion process, followed by $\mathcal{T}$ step denoising process with weather-robust 4D radar features as conditions to get the denoised clear LiDAR features.}
  \label{fusion}
\end{figure*}

\subsection{Adverse Weather Simulation}
To analyze the performance under adverse weather conditions on our V2X-R dataset, we applied fog \cite{fogsim} and snow \cite{snowsim} simulations to the LiDAR point clouds based on physical reflection and geometric optics method. Recent research \cite{robustbench, MVDNet} has confirmed the consistency of perception performance between simulated and real-world adverse weather data. Detailed implementations of fog and snow simulations will be introduced in the supplementary material.
 


%% file: sec/4_Method.tex
\section{Cooperative LiDAR-4D Radar Fusion}
\subsection{Overall Statement}
Built upon the V2X-R dataset, we present the first exploration of cooperative LiDAR-4D radar fusion for 3D object detection. To the best of our knowledge, no prior work has investigated this problem. 
In sec. \ref{sec:fusion_pipeline}, we introduce the first cooperative LiDAR-4D radar fusion pipeline for weather-robust 3D object detection.
Furthermore, in sec. \ref{sec:mdd}, we enhance the weather-robustness of cooperative LiDAR-4D radar fusion by leveraging the advantage of 4D radar in adverse weather conditions. 
We analyze the strengths and challenges of cooperative LiDAR point clouds in adverse weather conditions and propose the Multi-modal Denoising Diffusion (MDD) module to tackle these challenges, significantly improving the pipeline's weather-robustness.

\subsection{Fusion Pipeline}
\label{sec:fusion_pipeline}
In the cooperative LiDAR-4D radar fusion pipeline, there are three types of agents: Ego Vehicle (E), Connected Automated Vehicle (C), and Infrastructure (I). Each agent collects LiDAR and 4D radar point cloud data, forming the multi-agent multi-modal input $\mathbf{X} = \{X^L_C,X^L_E,X^L_I,X^R_C,X^R_E,X^R_I\}$. As illustrated in Fig. \ref{fusion}(a), subsequent fusion consists of four stages:

\textit{1) Encode by agent.} We feed $\mathbf{X}$ into a agent-shared encoder $\mathcal{G}_{\theta}$ to obtain features for each agent and modality as:
\begin{equation}
    \mathcal{F}^{i}_{j} = \mathcal{G}_{\theta}^i(X^{i}_{j}),
\end{equation}
where $i \in \{L,R\}$ denotes modality and $j \in \{C,E,I\}$ denotes agent.

\textit{2) Agent fusion.} After encoding, the features with same modality in different agent \{$\mathbf{\mathcal{F}}^{i}_{C}$, $\mathbf{\mathcal{F}}^{i}_{E}$, $\mathbf{\mathcal{F}}^{i}_{I} | i \in \{L,R\}$ communicate with each other, the agent-fusion $\phi_{\mathcal{A}}$ is performed to get multi-agent feature $\mathcal{F}^{i}_{\mathcal{A}}$ for $i$ modality  as:
\begin{equation}
    \mathcal{F}^{i}_{\mathcal{A}} = \phi_{\mathcal{A}}(\mathbf{\mathcal{F}}^{i}_{C}, \mathbf{\mathcal{F}}^{i}_{E}, \mathbf{\mathcal{F}}^{i}_{I}).
\end{equation}

\textit{3) Modal fusion.} The weather-induced noisy LiDAR feature $\mathcal{F}^{L}_{\mathcal{A}}$ will first be denoised to clear LiDAR feature $\tilde{\mathcal{F}^{L}_{\mathcal{A}}}$ by MDD module (will be described in the next section). Then, we perform modal-fusion $\phi_{\mathcal{M}}$ to obtain the multi-agent multi-modal features $\mathbf{\mathcal{F}}^{\mathcal{M}}_\mathcal{A}$ as:
\begin{equation}
    \mathbf{\mathcal{F}}^{\mathcal{M}}_\mathcal{A} = \phi_{\mathcal{M}}(\tilde{\mathcal{F}^{L}_{\mathcal{A}}}, \mathbf{\mathcal{F}}^{R}_{\mathcal{A}}).
\end{equation}

\textit{4) Box prediction.} Finally, we use the detector head to predict the 3D bounding box $\mathbf{B}$ by using the multi-agent multi-modal features $\mathbf{\mathcal{F}}^{\mathcal{M}}_\mathcal{A}$.

The two keys to implementing the above fusion pipeline lie in the agent-fusion and modal-fusion stages. Since there is no cooperative LiDAR-4D radar fusion method available yet. We explored two implementations that extend the existing mature fusion to cooperative LiDAR-4D radar fusion: 

(a) \textbf{S}ingle-\textbf{A}gent Multi-Modal \textbf{to} \textbf{M}ulti-\textbf{A}gent Multi-Modal (\texttt{SA2MA}). LiDAR-4D radar fusion in single-agent has received considerable attention \citep{InterFusion, MMFusion, L4DR, ourl4dr}. Therefore, we extend existing single-agent LiDAR-4D radar fusion methods to the multi-agent version by integrating a self-attention-based agent-fusion module (2nd stage). 

(b) Multi-Agent \textbf{S}ingle-\textbf{M}odal \textbf{to} Multi-Agent \textbf{M}ulti-\textbf{M}odal (\texttt{SM2MM}). Another approach is to extend the widely researched multi-agent LiDAR-based methods \citep{where2comm,sicp,opv2v,cobevt,v2xvit} to the multi-modal version. Specifically, we first extract multi-agent features from LiDAR and 4D radar point clouds individually and then concatenate BEV features in multi-modal fusion (3rd stage).

\begin{figure}[!t]
  \includegraphics[width=1.0\columnwidth]{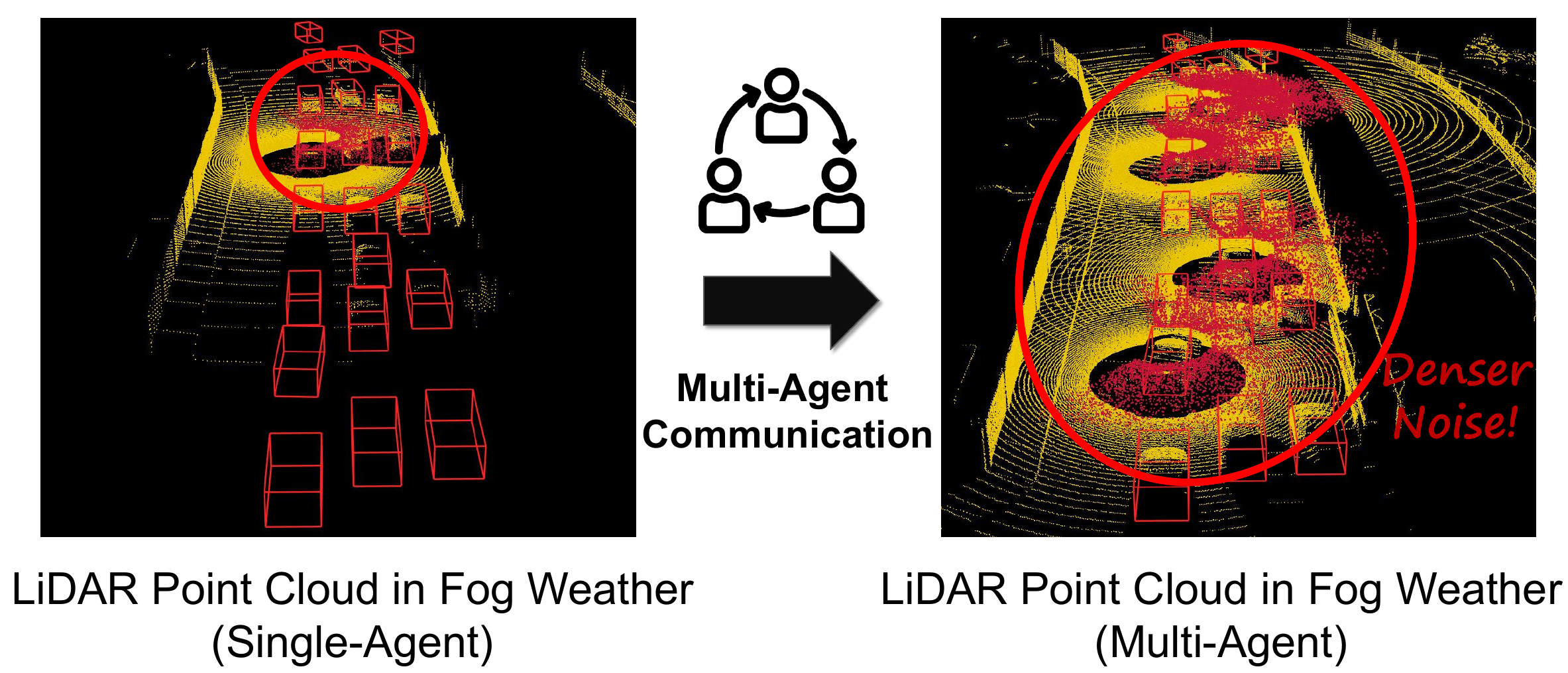}
  \caption{Visualization of LiDAR point cloud under foggy (simulated) weather before and after multi-agent communication. After multi-agent communication, the LiDAR point cloud has a longer vision, but a serious challenge of denser noise (red points).}
  \label{noise}
\end{figure}

\subsection{Multi-modal Denoising Diffusion (MDD) }
\label{sec:mdd}
We first analyzed the cooperative LiDAR point cloud in adverse weather. As shown in Fig. \ref{noise}, single-agent LiDAR point clouds exhibit the effects of reduced detection range and weather noise in fog weather. With multi-agent communication, the effect of reduced detection range is largely solved, but at the same time, weather noises exacerbate as they propagate through communication.

Therefore, addressing the dense noise challenge is crucial for weather-robust 3D object detection. Traditional point-level segmentation for denoising is time-consuming and incompatible with common multi-agent feature fusion strategies. On the other hand, simple feature-level denoising leads to poor fitting due to the complex distribution of weather feature noise. To overcome these limitations, we propose the Multi-modal Denoising Diffusion (MDD) for denoising noisy LiDAR features. As shown in Fig. \ref{fusion}(b), we resampled the complex weather noise distribution to a Gaussian distribution through the diffusion process, and then denoised it with the weather-robust 4D radar feature.

Specifically, drawing inspiration from DDPM \citep{diffusion} and MTL \citep{diffusionmtl}, gaussian noise is gradually applied to the init feature $\mathcal{F}_{init}$ (Initialization is $\mathcal{F}^{L}_{\mathcal{A}}$) in $\mathcal{T}$ times step-by-step. Suppose the decayed feature at denoising step $t$ is $\mathcal{F}_t$, the diffusion process $q$ can be formulated as:
\begin{equation}
    q(\mathcal{F}_t|\mathcal{F}_{init}) = \mathcal{N}(\mathcal{F}_t|\sqrt{\overline{\alpha}_t}\mathcal{F}_{init}, (1-\overline{\alpha}_t)\mathbf{I}),
\end{equation}
where$\{\overline{\alpha}_t,t\in\{1,2,...,\mathcal{T}\}\}$ are hyper-parameters. We could directly compute the decayed feature $\mathcal{F}_\mathcal{T}$ through mathematical derivation:
\begin{equation}
    \mathcal{F}_\mathcal{T} = \sqrt{\overline{\alpha}_\mathcal{T}}\mathcal{F}_{init} + \sqrt{1-\overline{\alpha}_\mathcal{T}}\epsilon,  \epsilon \sim \mathcal{N}(0, \mathbf{I}).
\end{equation}

This process transforms the original noise distribution $\delta_{raw}$ into a Gaussian distribution $\delta_{gau}$ through a Gaussian reparameterization sampling process as:
\begin{equation}
    \delta_{gau} =  \sqrt{\overline{\alpha}_\mathcal{T}}\delta_{raw} + \sqrt{1-\overline{\alpha}_\mathcal{T}}\epsilon, 
\end{equation}
$ \delta_{gau} \sim     \mathcal{N}(\sqrt{\overline{\alpha}_\mathcal{T}}\delta_{raw}, \sqrt{1-\overline{\alpha}_\mathcal{T}})$ subject to Gaussian distribution, which facilitates the fitting of the denoising model. Subsequently, at each denoising step, we incorporate weather-robust 4D radar features $\mathcal{F}^R_\mathcal{A}$ as a condition to prompt U-Net~\cite{unet} denoiser $U_\theta$ to predict the denoised feature:
\begin{equation}
    \mathcal{F}_{t-1} = U_\theta([\mathcal{F}_t,\mathcal{F}^R_\mathcal{A}],t), t\in\mathcal{T}, \mathcal{T}-1,...,1.
\end{equation}

\begin{algorithm}[!t]
    
	\caption{Multi-modal Denoising Diffusion process} 
	\label{alg1} 
	\begin{algorithmic}
        
		\STATE \textbf{Input:} Training $\in$ \{True, False\}; Noisy LiDAR BEV feature $\mathcal{F}^L_\mathcal{A}$; Noise-masked LiDAR BEV feature $\mathcal{F}^L_l$; 4D radar BEV feature $\mathcal{F}^R_\mathcal{A}$;  Diffusion steps $\mathcal{T}$; 
        \STATE \textbf{Output:} Training loss and denoised LiDAR feature $\tilde{\mathcal{F}^{L}_{\mathcal{A}}}$
		\STATE $\mathcal{F}_{init} \gets \mathcal{F}^L_\mathcal{A}$
        \STATE $\epsilon \in \mathcal{N}(0,\mathbf{I})$ \COMMENT{Sample noise}
        \STATE $\mathcal{F}_\mathcal{T} = \sqrt{\overline{\alpha}_t}\mathcal{F}_{init} + \sqrt{1-\overline{\alpha}_\mathcal{T}}\epsilon $ \COMMENT{Diffusion process}
        \FOR{$t \gets \mathcal{T}, \mathcal{T}-1, ..., 1$}
        \STATE $\mathcal{F}_{t-1} = U_\theta(Concat[\mathcal{F}_t,\mathcal{F}^R_\mathcal{A}],t)$  \COMMENT{Denoising process}
        \ENDFOR
        \STATE $\tilde{\mathcal{F}^{L}_{\mathcal{A}}} \gets \mathcal{F}_0$
        \IF{Training} 
        \STATE Loss $\gets \mathcal{L}_{MDD}(\tilde{\mathcal{F}^{L}_{\mathcal{A}}}, \mathcal{F}^L_l)$  \COMMENT{Compute loss as in eqs. \ref{eq_mdd}}
        \RETURN Loss, $\tilde{\mathcal{F}^{L}_{\mathcal{A}}}$ \COMMENT{Get loss and denoised feature}
        \ELSE 
        \RETURN $\tilde{\mathcal{F}^{L}_{\mathcal{A}}}$ \COMMENT{Get denoised clear LiDAR feature}
        \ENDIF
        
	\end{algorithmic} 
\end{algorithm}

For the final output after $\mathcal{T}$ denoising steps, the $U_\theta$ generates the denoised clear LiDAR feature $\tilde{\mathcal{F}^{L}_{\mathcal{A}}} = \mathcal{F}_0$. We present the detailed training and inference pipelines  MMD in Algorithm \ref{alg1}. And we compute the loss of MDD  as :
\begin{equation}
\label{eq_mdd}
\mathcal{L}_{MDD} =  \mathcal{L}_{MSE}(\tilde{\mathcal{F}^{L}_{\mathcal{A}}},\mathcal{F}^L_l) * \gamma(e, \psi),
\end{equation}
    where $\mathcal{F}^L_l$ is the groundtruth feature extracted from the clear LiDAR point cloud after masking the weather noise, $e$ is the epoch, $\psi$ is loss weight and $\gamma(e)$ is the loss weight that we specifically designed for the MDD module as:
\begin{equation}
\label{weight}
\gamma(e,\psi) = (1-tanh(\frac{e}{\tau} - \varphi))* \psi,
\end{equation}
where $\tau$ is temperature, $\varphi$ is offset. It decreases nonlinearly with epoch so that the model pays full attention to the feature denoising task in the early period and the object detection task in the late period.

\subsection{Loss Function}
We trained models with our MDD by the following losses:
\begin{equation}
\label{eq_all}
\mathcal{L}_{all} = \beta_{cls}\mathcal{L}_{cls}+\beta_{loc} \mathcal{L}_{loc}+ \mathcal{L}_{MDD},
\end{equation}
where $\beta_{cls}$, $\beta_{loc}$ are hyper-parameters, $\mathcal{L}_{cls}$ and  $\mathcal{L}_{loc}$ are classification and localization loss, respectively. All hyper-parameters are detailed in the supplementary material.

%% file: sec/5_Experiments.tex
\section{Experiments}
\subsection{Experimental Details and Metrics}
\definecolor{LightCyan}{rgb}{0.95,0.95,0.95}
We used 8,084/829/3,166 frames for training/ validation/ testing in our V2X-R dataset, ensuring there is no overlap in the intersection of the training/validation/testing sets. For each frame, we ensure that the minimum and maximum numbers of agents are 2 and 5 respectively. We use Adam optimizer with lr = 1e-3, $\beta_1$ = 0.9, $\beta_2$ = 0.999.
Following \citep{opv2v}, we select a vehicle as the ego vehicle for evaluation. Detection performance is evaluated near the ego vehicle in a range of x $\in$ [0, 140]m, y $\in$ [-40, 40]m. We set the broadcast range among CAVs to be 70 meters. In addition, since the 4D radar sensor only provides front-view data, all of our evaluation results are in the camera FOV of the ego vehicle.
\subsection{Benchmark Models}
We implement various state-of-the-art 3D object detectors on the V2X-R dataset, including different numbers of agents and different modalities. The results of these detectors on three different modalities are also given. The chosen 3D detector models for different modalities are as follows.

\textbf{Cooperative LiDAR-based 3D object detectors.} We pick existing method including AttFuse \citep{opv2v}, V2XViT \citep{v2xvit}, CoBEVT \cite{cobevt}, CoAlign \cite{coalign}, Where2comm \cite{where2comm}, AdaFusion \cite{adafusion}, SCOPE \cite{scope}, and SICP \cite{sicp} for cooperative LiDAR-based benchmarking analysis.

\label{sec:5.2}
\textbf{Cooperative 4D radar-based 3D object detectors.} 
Due to the lack of cooperative 4D radar-based methods. We first extend the single-agent 4D radar-based FPA-Net \citep{FPA-Net} and RTNH \citep{K-radar} to the cooperative 4D radar-based method through self-attention agent fusion in AttFuse \citep{opv2v}, marked as $\dagger$. For a more comprehensive analysis, we also changed the LiDAR point cloud input for the above cooperative LiDAR-based method to 4D radar point cloud input to implement cooperative 4D radar-based methods, marked as $\ddagger$.

\begin{table}[!t]
\resizebox{\columnwidth}{!}{%
\begin{tabular}{c|c|c|c}
\bottomrule
\textbf{Method} & \textbf{Publication} & \textbf{\begin{tabular}[c]{@{}c@{}}3D mAP@Validation\\ (IoU=0.3/0.5/0.7)\end{tabular}} & \textbf{\begin{tabular}[c]{@{}c@{}}3D mAP@Testing\\ (IoU=0.3/0.5/0.7)\end{tabular}} \\ \hline
V2XViT \citep{v2xvit}                                                                        & ECCV2022 &83.47/80.65/63.48&89.44/88.40/77.13\\
Attfuse \citep{opv2v}& ICRA2022&86.69/82.58/66.56&91.21/89.51/80.01\\
Where2comm \citep{where2comm}&NeurIPS2023&85.31/82.65/64.35&85.59/84.27/73.13\\
SCOPE \citep{scope}& ICCV2023& 79.43/77.35/65.08  & 80.67/79.88/72.00
\\
CoBEVT  \cite{cobevt}                                                                      & CoRL2023&86.65/84.59/70.30&91.41/90.44/81.06\\
CoAlign  \cite{coalign}                                                                     & ICRA2023&84.43/82.29/70.68&88.12/86.99/80.05\\
AdaFusion\cite{adafusion}&WACV2023&88.19/86.96/75.55&92.72/91.64/84.81\\
SICP    \cite{sicp}& IROS2024 &81.08/77.56/58.10&84.65/82.18/66.73\\ \toprule
\end{tabular}%
}
\caption{Experimental 3D object detection results of various cooperative LiDAR-based methods on the validation and testing of our V2X-R dataset in different IoU (0.3,0.5,0.7).}
\label{tab:LiDAR-based (Multi-Agent) Benchmark}
\end{table}

\begin{table}[!t]
\resizebox{\columnwidth}{!}{%
\begin{tabular}{c|c|c|c}
\bottomrule
\textbf{Method} & \textbf{Publication} & \textbf{\begin{tabular}[c]{@{}c@{}}3D mAP@Validation\\ (IoU=0.3/0.5/0.7)\end{tabular}} & \textbf{\begin{tabular}[c]{@{}c@{}}3D mAP@Testing\\ (IoU=0.3/0.5/0.7)\end{tabular}} \\ \hline
PFA-Net$^\dagger$ \cite{RPFA-Net}                                                           & ITSC2021             & 75.45/66.66/38.30&84.93/79.71/52.46	\\
RTNH$^\dagger$     \cite{RTNH}                                                          & NeurIPS2022          &72.00/62.54/34.65&73.61/67.63/41.86\\

V2XViT$^\ddagger$     \cite{v2xvit}                                                                     & ECCV2022 &68.58/61.88/29.47&80.61/73.52/42.60\\
AttFuse$^\ddagger$      \cite{opv2v}                                                        & ICRA2022             & 71.83/63.33/34.15&81.34/74.98/47.96 \\
Where2comm$^\ddagger$  \cite{where2comm}       & NeurIPS2023 &68.99/56.55/25.59&79.21/72.88/36.15\\
SCOPE$^\ddagger$\cite{scope} & ICCV2023 & 71.97/67.03/53.23& 72.20/69.12/53.41\\
CoBEVT$^\ddagger$     \cite{cobevt}                                                                      & CoRL2023 & 73.48/66.82/34.48&85.74/80.64/54.34  \\

CoAlign$^\ddagger$  \cite{coalign}                                                                      & ICRA2023             &75.05/68.11/41.20&81.69/75.74/52.01	\\
AdaFusion$^\ddagger$     \cite{adafusion}                                                                  & WACV2023             & 75.60/70.33/41.11&81.95/77.84/55.32 \\

SICP$^\ddagger$     \cite{sicp}                                                                       & IROS2024             & 70.83/62.79/34.82&71.94/65.17/36.44\\
\toprule
\end{tabular}%
}
\caption{Experimental 3D object detection results of various cooperative 4D Radar-based methods on the validation and testing of our V2X-R dataset in different IoU (0.3,0.5,0.7). The representations of $\dagger$ and $\ddagger$ as described in \ref{sec:5.2}.}
\label{tab:4DRadar-based (Multi-Agent) Benchmark}
\end{table}

\begin{table*}[!t]
\centering
\resizebox{\textwidth}{!}{%
\begin{tabular}{ccc|ccccccc}

\bottomrule
\cellcolor[HTML]{FFFFFF}                                                                                                     & \cellcolor[HTML]{FFFFFF}                                       & \cellcolor[HTML]{FFFFFF}                                       & \multicolumn{3}{c}{\cellcolor[HTML]{FFFFFF}\textbf{3D mAP@Validation}} & \multicolumn{3}{c}{\cellcolor[HTML]{FFFFFF}\textbf{3D mAP@Testing}} \\ 
\multirow{-2}{*}{\cellcolor[HTML]{FFFFFF}\textbf{Method}} & \multirow{-2}{*}{\cellcolor[HTML]{FFFFFF}\textbf{Publication}} & \multirow{-2}{*}{\cellcolor[HTML]{FFFFFF}\textbf{Fusion Strategy}} & \textbf{IoU=0.3}    & \textbf{IoU=0.5}    & \textbf{IoU=0.7}    & \textbf{IoU=0.3}   & \textbf{IoU=0.5}   & \textbf{IoU=0.7}   \\ \hline
InterFusion \cite{InterFusion}& IROS2022& \texttt{SA2MA}&78.33&74.70&51.44& 87.91& 86.51& 69.63              \\
L4DR \cite{ourl4dr} & AAAI2025&\texttt{SA2MA}&80.91&79.00&67.17&90.01&88.85&82.26              \\ \hline\hline
V2XViT \cite{v2xvit}& ECCV2022 &\texttt{SM2MM}&85.43&83.32&66.23&91.21&90.07&79.87\\
AttFuse \cite{opv2v}& ICRA2022& \texttt{SM2MM}&83.45&81.47&69.11&91.50&90.04&82.44\\
SCOPE \cite{scope}& ICCV2023& \texttt{SM2MM}&79.84&79.10&65.31&84.14&83.61&73.15\\
Where2comm  \cite{where2comm}& NeurIPS2023& \texttt{SM2MM}&88.05&85.98&69.94&92.20&91.16&81.40\\
CoBEVT \cite{cobevt} & CoRL2023& \texttt{SM2MM}&86.45&85.49&\textbf{75.65}&\textbf{94.23}&\textbf{93.50}&\textbf{86.92}	\\
CoAlign \cite{coalign}& ICRA2023& \texttt{SM2MM}&87.08&85.44&73.66&91.13&90.19&83.73\\
AdaFusion \cite{adafusion}& WACV2023 & \texttt{SM2MM} &\textbf{88.87}&\textbf{86.94}&74.44&92.94&91.97&85.31\\
SICP \cite{sicp}& IROS2024 & \texttt{SM2MM} &83.32&80.61&63.08&84.83&82.59&67.61 \\
\toprule
\end{tabular}%
}
\caption{Results of various cooperative LiDAR-4D radar fusion methods on the validation and testing of our V2X-R dataset in different IoUs (0.3,0.5,0.7). \texttt{SA2MA} and \texttt{SM2MM} represent \textit{Single-Agent to Multi-Agent} and \textit{Single-Modal to Multi-Modal} as described in sec. \ref{sec:fusion_pipeline}.}
\label{tab:LiDAR-4DRadar Fusion (Multi-Agent) Benchmark}
\end{table*}

\begin{table*}[!t]
\resizebox{\textwidth}{!}{%
\begin{tabular}{cc|ccccccccc}

\bottomrule
                                 & & \multicolumn{3}{c}{\textbf{3D mAP@Snow}}       & \multicolumn{3}{c}{\textbf{3D mAP@Fog}}       & \multicolumn{3}{c}{\textbf{3D mAP@Normal}}      \\

\multirow{-2}{*}{\textbf{Method}} &\multirow{-2}{*}{\textbf{Modality}} & IoU=0.3        & IoU=0.5        & IoU=0.7        & IoU=0.3        & IoU=0.5        & IoU=0.7        & IoU=0.3        & IoU=0.5        & IoU=0.7        \\ \hline

\rowcolor[HTML]{FFFFFF} 
L4DR \cite{ourl4dr}     &L+4DR                        & 78.88          & 75.87          & 59.10          & 85.00          & 80.09          & 61.48          & \textbf{90.66} & \textbf{89.29} & \textbf{80.42} \\
L4DR w/MDD   &L+4DR                    & \textbf{82.89} & \textbf{80.52} & \textbf{65.80} & \textbf{86.00} & \textbf{81.60} & \textbf{63.66} & 90.44          & 89.15          & 80.36          \\
\rowcolor{LightCyan}
\multicolumn{2}{c|}{\textit{Improvement}}              & \textit{+4.01} & \textit{+4.65} & \textit{+6.70} & \textit{+1.00} & \textit{+1.51} & \textit{+2.18} & \textit{-0.22} & \textit{-0.14} & \textit{-0.06} \\

\hline
\hline
\rowcolor[HTML]{FFFFFF} 
1.AttFuse  \cite{opv2v} &L                           & 68.73          & 65.35          & 45.54          & 81.71          &78.48           &61.52          & 89.30 & 87.39 & 76.09           \\
2.AttFuse \cite{opv2v}  &L+4DR                           & 79.63          & 77.37          & 63.71          & 85.00          & 80.64          & 62.89          & \textbf{91.15} & \textbf{89.80} & 81.75          \\
\rowcolor[HTML]{FFFFFF} 
3.AttFuse w/MDD   &L+4DR                   & \textbf{83.78} & \textbf{81.19} & \textbf{66.86} & \textbf{87.37} & \textbf{83.90} & \textbf{68.64} & 90.99          & 89.64          & \textbf{81.96} \\
\rowcolor{LightCyan} 
\multicolumn{2}{c|}{\textit{Improvement (2-1)}}              & \textit{+10.90} & \textit{+12.02} & \textit{+18.17} & \textit{+3.29} & \textit{+2.16} & \textit{+1.37} & \textit{+1.85} & \textit{+2.41} & \textit{+5.66} \\
\rowcolor{LightCyan}
\multicolumn{2}{c|}{\textit{Improvement (3-2)}}              & \textit{+4.15} & \textit{+3.82} & \textit{+3.15} & \textit{+2.37} & \textit{+3.26} & \textit{+5.73} & \textit{-0.16} & \textit{-0.16} & \textit{+0.21} \\

\toprule
\end{tabular}%
}
\caption{The 3D mAP performance comparison under different weather conditions on the V2X-R dataset. 'L' and '4DR' represent LiDAR and 4D radar, respectively. '-' indicates that multi-modal method L4DR \cite{ourl4dr} cannot achieve LiDAR-based performance.}
\label{tab:Multi-modal Diffusion Feature Denoise (MDFD)}
\end{table*}

\textbf{Cooperative LiDAR-4D radar fusion 3D object detectors.} We implement \texttt{SA2MA} fusion on the InterFusion \citep{InterFusion} and L4DR \citep{ourl4dr} models and \texttt{SM2MM} fusion on the AttFuse \citep{opv2v}, CoBEVT \citep{opv2v}, V2XViT\citep{opv2v}, Where2comm \citep{where2comm}, CoAlign \citep{coalign}, AdaFusion \citep{adafusion}, SICP \cite{sicp} models. 

To intuitively compare the performance in normal weather differences between modalities, the cooperative LiDAR-4D radar fusion 3D object detectors we implemented in the benchmark section do not incorporate the MDD module.
The MDD module will be discussed in subsequent experiments.

\subsection{Benchmark Analysis}
In the following, we will show and analyze the results of three different modalities on our V2X-R dataset.

\textbf{Cooperative LiDAR-based 3D object detectors.}
Table \ref{tab:LiDAR-based (Multi-Agent) Benchmark} depicts the performance of the selected cooperative LiDAR-based 3D object detectors. A large number of combined results from these models validate the reliability of our V2X-R dataset. It is worth noting that the performance of the newest methods (SICP \cite{sicp}) is not as good as that of earlier methods (e.g. COBEVT \citep{cobevt}, CoAlign \citep{coalign}, AdaFusion \citep{adafusion}). This is because SICP \cite{sicp} not only focus on mainstream performance but also emphasize preserving the ability of individual perception.

\textbf{Cooperative 4D radar-based 3D object detectors.}
We also pioneered the exploration of cooperative 4D radar-based methods on the V2X dataset. As shown in Table \ref{tab:4DRadar-based (Multi-Agent) Benchmark}, the performance of 4D radar-based models is generally lower than that of LiDAR-based models in Table \ref{tab:LiDAR-based (Multi-Agent) Benchmark}. This can be attributed to the significantly lower resolution of 4D radar than LiDAR, which is a limitation hindering the independent use of 4D radar in single-agent scenarios. However, we also observed that cooperative 4D radar-based methods can achieve the excellent performance of 85.7\% 3D mAP@testing (IoU=0.3) (CoBEVT$^\ddagger$ \citep{cobevt}). This shows that 4D radar can achieve independent perception through cooperative communication, proving the research significance of the cooperative 4D radar-based approach.


\textbf{Cooperative LiDAR-4D radar fusion 3D object detectors.}
Finally, we implemented and evaluated a series of cooperative LiDAR-4D radar fusion methods following the methodology described in sec. \ref{sec:fusion_pipeline}. As shown in Table \ref{tab:LiDAR-4DRadar Fusion (Multi-Agent) Benchmark}, by comparing with Tables \ref{tab:LiDAR-based (Multi-Agent) Benchmark} and \ref{tab:4DRadar-based (Multi-Agent) Benchmark}, it can be observed that almost all cooperative LiDAR-4D radar fusion models achieve the best performance compared to other single-modality approaches. Especially, CoAlign \citep{coalign}, CoBEVT \cite{cobevt}, and Where2comm \citep{where2comm} exhibit 3.68\%, 5.86\%, and 8.23\% 3D mAP@testing (IoU=0.7) improvements through LiDAR-4D radar fusion, respectively. These results convincingly validate that cooperative 4D radar can bring considerable performance enhancements to the LiDAR single-modality approach. On the other hand, the table also shows that there is a significant difference in the benefits of adding 4DRadar across different methods, which indicates that exploring how to maximize the value of cooperative LiDAR-4D Radar fusion is still worth exploring.

\subsection{Multi-modal Diffusion Denoising Analysis}
To explore the weather-robust advantages of 4D radar, we investigated the performance of the LiDAR-4D radar fusion model and verified the effectiveness of our designed MDD module. We trained models by randomly sampling LiDAR point clouds in [normal, fog] for each frame to test on various weather domains: normal weather domain, seen (fog) and unseen (snow) adverse weather domain.

\textbf{Performance comparison under different simulated weather on V2X-R dataset.} We selected three models, AttFuse \cite{opv2v} and L4DR \cite{ourl4dr}, which include \texttt{SA2MA} and \texttt{SM2MM} fusion strategies, to evaluate the performance under different weather conditions. As shown in Table \ref{tab:Multi-modal Diffusion Feature Denoise (MDFD)}, the LiDAR-4D radar fusion can achieve superior performance in various weather conditions. AttFuse achieved 18.17\% 3D mAP@snow (IoU=0.7) and 5.66\% 3D mAP@normal (IoU=0.7) improvement when compared to LiDAR-only (\textit{Improvement (2-1)}), respectively. Moreover, our designed MDD module improved the basic LiDAR-4D radar fusion model performance by up to 5.73\% (AttFuse, 3D mAP@fog) and 6.70\% (L4DR, 3D mAP@snow) under adverse weather conditions. Meanwhile, our MDD module hardly affects the normal weather performance, which validates the effectiveness and reliability of our MDD module.

\begin{table}[!t]
\resizebox{\columnwidth}{!}{%
\begin{tabular}{c|c|c|cc}
\bottomrule
                                                &                                   &  &  & \\ 
\multirow{-2}{*}{\textbf{Class}}                & \multirow{-2}{*}{\textbf{Method}}  & \multirow{-2}{*}{\textbf{All}}        & \multirow{-2}{*}{\textbf{Adverse}}              & \multirow{-2}{*}{\textbf{Normal}}              \\ \hline             & AttFuse$^\dagger$ \cite{opv2v}                      & 69.04           & 70.72       & 66.75                \\
\multirow{-2}{*}{Sedan} & AttFuse$^\dagger$ w/ MDD              & \textbf{74.03} & \textbf{75.90} & \textbf{67.18}       \\
\hline  & AttFuse$^\dagger$ \cite{opv2v}                      & 50.95          & 48.78  & \textbf{53.33}       \\
\multirow{-2}{*}{\cellcolor[HTML]{FFFFFF}Bus}   & AttFuse$^\dagger$ w/ MDD              & \textbf{54.58} & \textbf{54.75}         & 52.81                \\ \toprule
\end{tabular}%
}
\caption{The 3D mAP performance comparison on K-Radar \cite{K-radar} dataset. Adverse represents average results under various adverse weather, including overcast, fog, rain, sleet, light and heavy snow. $^\dagger$ represents AttFuse~\cite{opv2v} is implemented as single agent method by removing the agent fusion module.}
\label{tab:CDFD module validation on KRadar dataset}
\end{table}

\textbf{Performance comparison under different real-world weather on K-Radar dataset.}
We further conducted experiments on the K-Radar single-agent real-world dataset. As shown in Table \ref{tab:CDFD module validation on KRadar dataset},  our MDD module demonstrates significant performance improvements in real-world adverse weather, with 3D mAP gains of 5.20\% and 5.97\% for the Sedan and Bus classes, respectively. MDD has rarely impacted the normal weather performance, resulting in overall improvements of 4.99\% and 3.63\%. These results validate the effectiveness of the MDD module in the presence of dense noise under real-world adverse weather.

\begin{table}[!t]
\resizebox{\columnwidth}{!}{%
\begin{tabular}{ccc|ccc}
\bottomrule
\multicolumn{3}{c|}{\textbf{Component}}  & \multicolumn{3}{c}{\textbf{3D mAP@Fog}}        \\
Vanilla-Unet & Diffusion & 4D radar & \multirow{2}{*}{IoU=0.3}        & \multirow{2}{*}{IoU=0.5}        & \multirow{2}{*}{IoU=0.7}        \\

 Denoise & Denoise &  Condition &&& \\
 \hline
 &&&  85.00          & 80.64          & 62.89          \\ 
\checkmark                                                              &                                                             &                                                                        & 85.44          & 81.23          & 62.07          \\ &   \checkmark                                                         &                                                               & 86.41          & 83.29          & 68.08          \\
                                                               &  \checkmark                                                        & \checkmark                                                             & \textbf{87.37} & \textbf{83.90} & \textbf{68.64} \\ \toprule
\end{tabular}%
}
\caption{Effect of each component in MDD module, tested by AttFuse \cite{opv2v} on V2X-R testing with fog-simulation.}
\label{tab:CDFD module ablation validation}
\end{table}

\textbf{Effect of each component.}
We evaluated the effect of each component, as shown in Table \ref{tab:CDFD module ablation validation}. We initially attempted to denoise LiDAR features using Vanilla-Unet (row 2$^{st}$) but found the results unsatisfactory. The complex distribution of LiDAR noise in multi-agent settings is difficult to directly fit. We significantly improve performance by transforming the noise distribution into Gaussian distribution through a multi-step diffusion process (row 3$^{rd}$). Finally, by incorporating weather-robust 4D Radar features as conditional information for multi-modal denoising (row 4$^{th}$), the model achieves the best performance of 87.37/83.90/86.64 3DmAP@Fog.

\textbf{Effect of diffusion layers and loss weight.}
Finally, we investigate the impact of hyper-parameters in our MDD module, including number of diffusion layers and the MDD loss weight $\mathcal{T}$ (both constant weights and the epoch-varying weight $\gamma(e,\psi)$ in eqs. \ref{weight}). As shown in Table \ref{tab:CDFD parametric experiments}, the best performance of 87.37/83.90/86.64 3DmAP@Fog is achieved with MDD loss weight $\gamma(e,\psi=3)$ and number of diffusion layers $\mathcal{T} = 3$. Although MDD inevitably introduces an additional inference time of 32 ms, it significantly improves weather robustness and still maintains real-time (about 20 FPS). These results validate the effectiveness of our designed weight $\gamma(e,\psi)$ and real-time of our MDD module.

\begin{table}[!t]
\resizebox{\columnwidth}{!}{%
\begin{tabular}{@{}c|c|c|ccc@{}}
\bottomrule      \textbf{Diffusion} &                           \textbf{Loss}               & \textbf{Inference}&\multicolumn{3}{c}{\textbf{3D mAP@Fog}}       \\
\textbf{ Layers} & \textbf{Weight} & \textbf{{Speed}} & IoU=0.3        & IoU=0.5        & IoU=0.7        \\ \hline
- &- & {18 ms}& 85.00 & 80.64 & 62.89 \\ \hline 
\multirow{6}{*}{$\mathcal{T} = 2$} & $1$ & \multirow{6}{*}{{39 ms}} & 85.85   & 82.19 & 65.47 \\
& $3$  & & 86.68  & 82.77  & 65.92\\
& $5$  & & 85.54  & 81.79          & 65.89 \\
& $\gamma(e,\psi = 1)$  & & 86.98          & 83.31          & 66.91          \\ 
& $\gamma(e,\psi = 3)$  & & 87.35          & 83.78          & 68.14          \\
& $\gamma(e,\psi = 5)$  & & 86.91          & 83.46          & 66.99          \\
\hline 
\multirow{2}{*}{$\mathcal{T} = 3$}    & $3$    &                           & 85.97          & 82.43          & 66.23          \\
& $\gamma(e,\psi = 3)$   & \multirow{-2}{*}{{50 ms}} & \textbf{87.37}  & \textbf{83.90}   & \textbf{68.64}          \\
\hline
\multirow{2}{*}{$\mathcal{T} = 4$}   & $3$   &    & 86.98        & 83.15          & 65.13 \\
& $\gamma(e,\psi = 3)$  &  \multirow{-2}{*}{{59 ms}}   & 87.01          & 83.89          & 67.58 \\ \toprule
\end{tabular}%
}
\caption{Effect of diffusion layers and loss weight in MDD module, tested by AttFuse \cite{opv2v} on V2X-R testing with fog-simulation.}
\label{tab:CDFD parametric experiments}
\end{table}

%% file: sec/6_Conclusion.tex
\section{Conclusion and Discussion}
We present V2X-R, the first simulated V2X collaborative perception dataset incorporating 4D radar. We establish a benchmark on V2X-R dataset by integrating a LiDAR-4D radar fusion pipeline into classic collaborative 3D object detection methods. Moreover, we propose the MDD module to tackle dense noise in collaborative conditions. In summary, V2X-R and our 
exploration paves the way for advancing collaborative LiDAR and 4D radar fusion perception.

\noindent \textbf{Limitations and future work.} 
While our work provides a foundation, deeper exploration of cooperative LiDAR and 4D radar fusion remains underdeveloped. A compelling research direction is the full utilization of multi-agent and multi-modal information for robust 3D object detection. 

\noindent  \textbf{Acknowledgements.} This work was supported in part by the National Natural Science Foundation of China (No.62171393), and the Fundamental Research Funds for the Central Universities (No.20720220064).

%% file: sec/X_suppl.tex
\maketitlesupplementary

\section{Additional Details of V2X-R Dataset}

\begin{figure}[!h]
\centering
  \includegraphics[width=0.7\columnwidth]{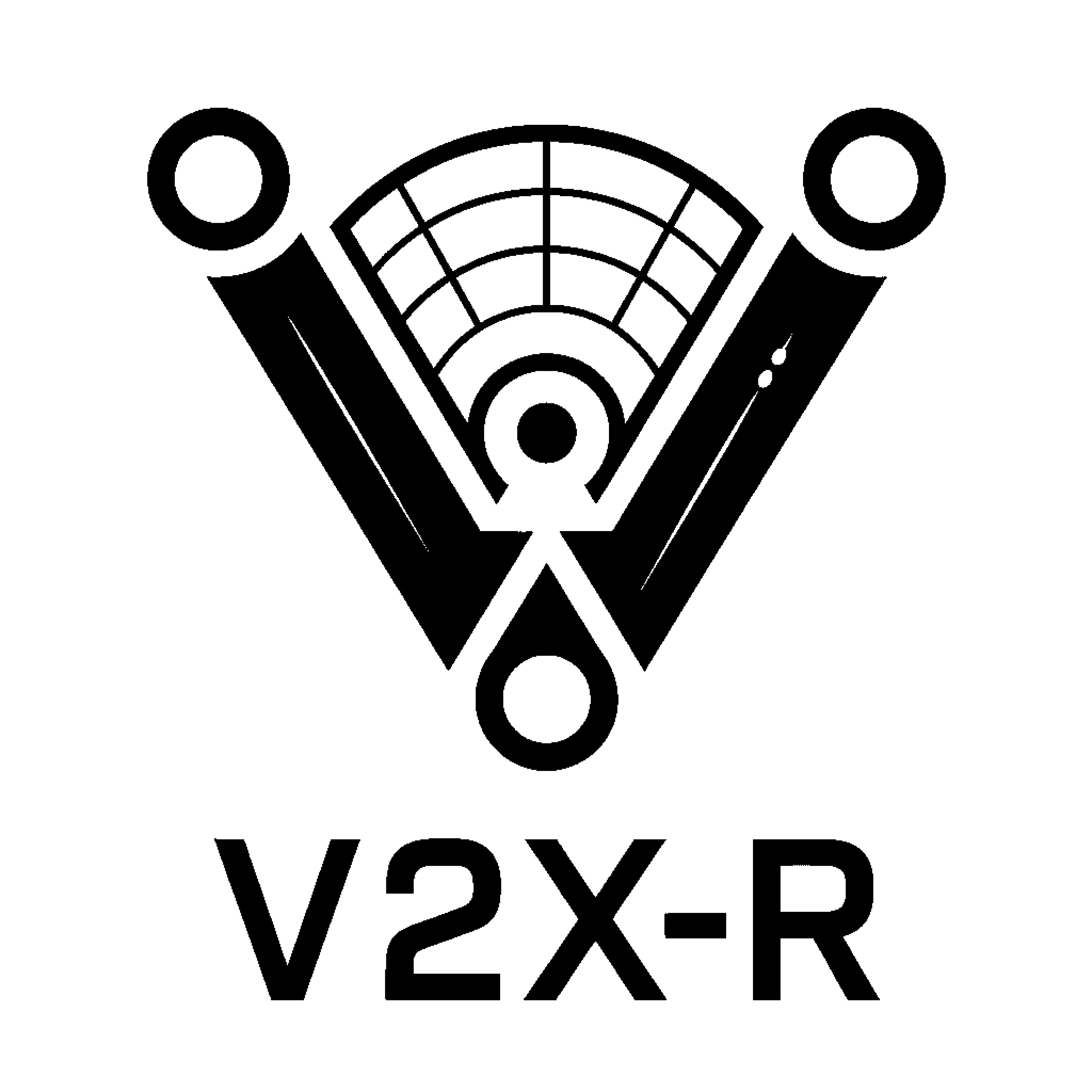}
  \label{t}
\end{figure}

Here we will introduce some additional details about our V2X-R dataset, to help researchers using the V2X-R dataset get started quickly.

\subsection{Calibration of sensors}
We provide calibration information for each sensor (LiDAR, 4D radar, camera) of each agent for inter-sensor fusion. In particular, the exported 4D radar point cloud has been converted to the LiDAR coordinate system of the corresponding agent in advance to facilitate fusion, so the 4D radar point cloud is referenced to the LiDAR coordinate system. If necessary, it can be converted back by the LiDAR coordinate system to the 4D radar coordinate system.

\begin{table}[!h]
\resizebox{\columnwidth}{!}{%
\begin{tabular}{c|c|c|c}
\bottomrule      \textbf{Sensor} &                           \textbf{Data Structure}  &             \textbf{File Type}   &\textbf{Attributes}       \\ \hline
\multirow{2}{*}{LiDAR} & \multirow{2}{*}{Point Cloud} & \multirow{2}{*}{.pcd} & $N\times4$ \\ 
& & & [$x,y,z,intensity$] \\ \hline
\multirow{2}{*}{4D Radar} & \multirow{2}{*}{Point Cloud} & \multirow{2}{*}{.pcd} & $N\times4$ \\ 
& & & [$x,y,z,velocity$] \\ \hline
\multirow{2}{*}{Camera} & \multirow{2}{*}{Image} & \multirow{2}{*}{.png} & $800\times600\times3$ \\ 
& & & [$R,G,B$] \\
\toprule

\end{tabular}%

}
\caption{The detailed information of V2X-R data.}
\label{info_data}
\end{table}

\subsection{Information of data}

The data attributes corresponding to LiDAR, 4D radar, and camera are shown in Table \ref{info_data}. Both LiDAR and 4D radar sensors provide $N\times4$ point clouds, where $N$ represents
the number of points. It is worth noting that the 4D radar was originally exported in the CARLA simulator as an array of [azimuth, altitude, depth, velocity] in polar coordinates, which we converted to a Cartesian coordinate system. In addition, an 800x600x3 RGB image is obtained for each of the 4 cameras of each agent.

\subsection{Information of data collection}

We saved the critical collection details of each scene sequence by splits (can be found in "data\_protocol.yaml" of the released dataset), which includes agent information and scene sources for each sequence. Additional detailed acquisition information, such as the configuration of each agent's trajectory, can be queried in our V2X-R documentation.

\section{Additional Results Analysis}

\begin{table}[!h]
\centering
\resizebox{\columnwidth}{!}{%
\begin{tabular}{
c |
c 
c 
c 
c 
c }
\hline
                       \multirow{2}{*}{Modality}    & \multicolumn{5}{c}{\begin{tabular}[c]{@{}c@{}}Communication volumes (4B as unit, log-scale)/\\ Latency time (ms as unit, 27Mbps as transmission speed)\end{tabular}} \\ \cline{2-6} 
 & AttFuse                            & V2XViT                            & CoBEVT                            & SICP                            & L4DR                                   \\ \hline
LiDAR-based                                        & 18.2/357.7 
                              & 16.0/75.2 
 
                             & 17.6/226.1 
                             & 14.9/35.2 
 
                           & -                                      \\
4D radar-based                                      &14.6/28.4 
                              & 14.0/18.6 
                             & 15.0/37.4 
                             & 13.6/14.7 
 
                           & -                                      \\
LiDAR+4D radar                                       & \cellcolor {gray!20}18.4/402.1 
                              &\cellcolor {gray!20}16.9/142.0 
                             &\cellcolor{gray!20}18.1/317.6 
                             &\cellcolor {gray!20}15.2/42.7 
                           & \cellcolor{green!10}17.7/253.1                           \\ \hline
\end{tabular}
}
\caption{Analysis of Communication Costs and Bandwidth for Different Models (\textcolor{gray}{SM2MM},\textcolor{green!50}{SA2MA}). }
\label{cost}
\end{table}

\subsection{Transmission cost and bandwidth.}

We have calculated transmission cost (count of non-zero elements in the feature map) and latency for different modalities. As shown in Table \ref{cost}, 4D radar has the advantage of sparse feature transmission, and its fusion with LiDAR brings acceptable transmission cost and latency time.

\begin{table}[!h]
\resizebox{\columnwidth}{!}{%
\begin{tabular}{
c |
c 
c 
c 
c 
c 
c 
c }
\hline
     & \multicolumn{7}{c}{Std of localization error   (m)/(Val-AP@50)}       \\ \cline{2-8} 
\multirow{-2}{*}{Method} & 0.0          & 0.1          & 0.2          & 0.3          & 0.4          & 0.5          & 0.6   \\ \hline
AttFuse                                                               & 84.30        & 81.28        & 79.87        & 77.84        & 73.09        & 68.30        & 62.84 \\
CoBEVT                                                                & 87.02        & 85.26        & 83.30        & 81.33        & 77.83        & 70.79        & 63.47 \\
AdaFusion                                                             & 87.31        & 87.15        & 86.25        & 84.22        & 79.79        & 72.89        & 64.15 \\
Where2comm                                                            & 85.78        & 85.58        & 84.78        & 82.91        & 79.57        & 76.23        & 71.77 \\ \hline
\end{tabular}
}
\caption{Performance of different models under varying degrees of localization error.}
\label{error}
\end{table}

\begin{table*}[!t]

\resizebox{\textwidth}{!}{%
\small
\begin{tabular}{cc|cccccccccc}
\hline
{{Methods}}                                           & {{Modality}}         & {{Class}} & {{Metric}} & {{Total}} & {{Normal}} & {{Overcast}} & {{Fog}} & {{Rain}} & {{Sleet}} & {{Lightsnow}} & {{Heavysnow}} \\ \hline\hline
\multirow{4}{*}{\shortstack{AttFuse \cite{opv2v}}}   & \multirow{4}{*}{L+4DR}
& \multirow{2}{*}{Sedan}  & $AP_{BEV}$ &70.3 &68.0 &\textbf{89.4} &90.5 &79.5 &66.9 &88.3 &60.4 \\
& &   & $AP_{3D}$ &69.0 &66.8 &79.4 &88.6 &70.7 &59.2 &\textbf{86.2} &58.6  \\ \cline{3-12}
&   &\multirow{2}{*}{Bus} & $AP_{BEV}$ &\textbf{64.3} &\textbf{59.4} &\textbf{75.7} &- &0.4 &\textbf{66.2} &80.7& 70.8 \\ 
& &  & $AP_{3D}$ &51.0 &\textbf{53.3 }&\textbf{75.6} &- &0.2 &\textbf{65.6} &76.8 &36.3  \\ \hline
\multirow{4}{*}{\shortstack{AttFuse w/ MDD}}   & \multirow{4}{*}{L+4DR}
& \multirow{2}{*}{Sedan}  & $AP_{BEV}$ &\textbf{76.8} &\textbf{73.8} & 88.9 &\textbf{90.8} &\textbf{79.7}& \textbf{68.7}&\textbf{88.4}&\textbf{61.5} \\
& &                     & $AP_{3D}$  &\textbf{74.0} &\textbf{67.2}&\textbf{85.5}&\textbf{89.6}&\textbf{75.7}&\textbf{64.6}&84.5&\textbf{59.8} \\ \cline{3-12}
& &\multirow{2}{*}{Bus} & $AP_{BEV}$ &64.1&55.3&72.0&-&\textbf{15.1}&62.7&\textbf{97.5}&\textbf{73.9} \\ 
& &                     & $AP_{3D}$  &\textbf{54.6} &52.8&71.0&-&\textbf{15.0}&61.3&\textbf{85.2}&\textbf{42.6}\\ \hline
\end{tabular}}
\caption{Quantitative results of different 3D object detection methods on K-Radar dataset. We present the modality of each method (L+4DR: LiDAR-4D radar fusion) and detailed performance for each weather condition. Best in \textbf{bold}.}
\label{full_kradar}

 \end{table*}

\begin{table*}[!t]
\centering
\resizebox{\textwidth}{!}{%
\begin{tabular}{
c |
c |
c |
c |
c |
c |
c }
\hline
\multirow{2}{*}{\textbf{Modality}} & \multirow{2}{*}{\textbf{Method}} & \multirow{2}{*}{\textbf{Epoch}} & \multirow{2}{*}{\textbf{Batch\_size}} & \multirow{2}{*}{\textbf{Max\_Agents}} & \multirow{2}{*}{\textbf{Learning\_Rate}} & \multirow{2}{*}{\textbf{LR\_Scheduler}} \\ &&&&&&\\ \hline                                          & V2XViT \cite{v2xvit}           & 20             & 2                    & 5                 & 0.001       & Multistep              \\ \cline{2-7} 
                                          & AttFuse \cite{opv2v}   & 30             & 4                    & 5                 & 0.002       & Multistep              \\ \cline{2-7} 
                                          & Where2comm \cite{where2comm}      & 50             & 1                    & 5                 & 0.0002      & Cosineannealwarm       \\ \cline{2-7} 
                                          & SCOPE \cite{scope}            & 30             & 2                    & 5                 & 0.002       & Multistep              \\ \cline{2-7} 
                                          & CoBEVT \cite{cobevt}          & 30             & 2                    & 5                 & 0.001       & Cosineannealwarm       \\ \cline{2-7} 
                                          & CoAlign \cite{coalign}         & 15             & 2                    & 5                 & 0.002       & Multistep              \\ \cline{2-7} 
                                          & AdaFusion \cite{adafusion}       & 30             & 2                    & 5                 & 0.0005      & Multistep              \\ \cline{2-7} 
                                          & SICP  \cite{sicp}           & 20             & 1                    & 5                 & 0.001       & Multistep              \\ \cline{2-7} 
\multirow{-9}{*}{\textbf{LiDAR}}          & MACP \cite{macp}            & 20             & 4                    & 5                 & 0.0002      & Cosineannealwarm       \\ \hline
                                          & PFA-Net  \cite{RPFA-Net}        & 30             & 4                    & 5                 & 0.001       & Multistep              \\ \cline{2-7} 
                                          & RTNH   \cite{RTNH}          & 15             & 4                    & 5                 & 0.001       & Multistep              \\ \cline{2-7} 
                                          & V2XViT  \cite{v2xvit}         & 20             & 2                    & 5                 & 0.001       & Multistep              \\ \cline{2-7} 
                                          & AttFuse \cite{opv2v}   & 30             & 4                    & 5                 & 0.002       & Multistep              \\ \cline{2-7} 
                                          & Where2comm \cite{where2comm}      & 15             & 1                    & 5                 & 0.0002      & Cosineannealwarm       \\ \cline{2-7} 
                                          & SCOPE \cite{scope}            & 15             & 2                    & 5                 & 0.002       & Multistep              \\ \cline{2-7} 
                                          & CoBEVT \cite{cobevt}          & 30             & 2                    & 5                 & 0.001       & Cosineannealwarm       \\ \cline{2-7} 
                                          & CoAlign  \cite{coalign}        & 20             & 2                    & 5                 & 0.002       & Multistep              \\ \cline{2-7} 
                                          & AdaFusion  \cite{adafusion}      & 15             & 2                    & 5                 & 0.0005      & Multistep              \\ \cline{2-7} 
\multirow{-10}{*}{\textbf{4D Radar}}       & SICP \cite{sicp}            & 20             & 1                    & 5                 & 0.001       & Multistep              \\ \hline
                                          & InterFusion  \cite{InterFusion}    & 20             & 1                    & 5                 & 0.002       & Multistep              \\ \cline{2-7} 
                                          & L4DR \cite{ourl4dr}            & 30             & 2                    & 5                 & 0.002       & Multistep              \\ \cline{2-7} 
                                          & V2XViT  \cite{v2xvit}         & 30             & 2                    & 5                 & 0.001       & Multistep              \\ \cline{2-7} 
                                          & AttFuse \cite{opv2v}  & 30             & 2                    & 5                 & 0.002       & Multistep              \\ \cline{2-7} 
                                          & SCOPE  \cite{scope}          & 40             & 2                    & 5                 & 0.002       & Multistep              \\ \cline{2-7} 
                                          & Where2comm \cite{where2comm}      & 30             & 4                    & 5                 & 0.0002      & Cosineannealwarm       \\ \cline{2-7} 
                                          & CoBEVT \cite{cobevt}          & 40             & 2                    & 5                 & 0.001       & Cosineannealwarm       \\ \cline{2-7} 
                                          & CoAlign  \cite{coalign}        & 30             & 2                    & 5                 & 0.002       & Multistep              \\ \cline{2-7} 
                                          & AdaFusion \cite{adafusion}       & 40             & 2                    & 5                 & 0.0005      & Multistep              \\ \cline{2-7} 
\multirow{-10}{*}{\textbf{LiDAR+4D Radar}} & SICP \cite{sicp}             & 20             & 1                    & 5                 & 0.001       & Multistep              \\ \hline
\end{tabular}%
}

\caption{ Experimental parameter settings (epoch, batch\_size, max\_agent, learing\_rate, lr\_scheduler) for different modalities and methods in our benchmark section.}
\label{Experimental parameter setting}
\end{table*}

\subsection{localization noise}

Following Where2comm \cite{where2comm}, we added different degrees of localization error to our V2X-R dataset to conduct experiments. As shown in Table \ref{error}, all methods experience performance degradation to varying degrees as localization error increases. This helps to explore performance under real localization errors.

\subsection{Performance of various weather on K-Radar.}
To further demonstrate the performance improvement of our MDD module in various real-world adverse weather conditions. We provide more detailed results on the K-Radar real adverse weather dataset rather than just the average of adverse weather. As shown in Table \ref{full_kradar}, with the addition of our MDD module, AttFuse first of all got a big boost in Total basically (except for Bus's $AP_{BEV}$). Under the Sedan class, there are significant improvements in every weather except Overcast which is similar, especially in 6.1 $AP_{3D}$@Overcast and 5.4 $AP_{3D}$@Sleet performance improvements. In addition, the results of the Bus class are well worth exploring. We find significant decreases in Normal, Overcast, and Sleet, but very significant increases in Rain, Lightsnow, and Heavysnow. We assert this is due to the nature of the larger 3D bounding boxes of the Bus class, which is particularly sensitive to the denoising module, causing a drop in some weather and a significant rise in others. Overall, however, the Total performance on the Bus class remains suggestive. These detailed analyses further validate the effectiveness of our MDD module under real-world conditions.

\begin{table}[!t]
\centering
\resizebox{\columnwidth}{!}{%
\begin{tabular}{c|c|c}
\bottomrule      \textbf{Component} &                           \textbf{Parameter}  &             \textbf{Value}   \\ \hline
\multirow{7}{*}{Denoiser (U-net)} &input\_channel &128\\
&mid\_channel &128\\
&timestep\_channel&64\\
&output\_channel&64\\
&number\_layers&2\\
&number\_resblock&2\\
 \hline\hline
\multirow{2}{*}{Diffusion Process} &timesteps &3\\
&betas &[0.005,0.0275,0.05]\\
\toprule

\end{tabular}%
}
\caption{The implementation details of our MDD module.}
\label{info_mdd}
\end{table}

\section{Training Detail}

\subsection{Benchmark}

We also provide details about the training of all the benchmark models in the main text for researchers to refer to, as shown in Table \ref{Experimental parameter setting}. In addition, we will disclose the training profiles of all models and the pre-trained models. This can help researchers efficiently use our well-trained models on the V2X-R dataset or reproduce the same results.

\subsection{Implementation of MDD }
Here, we will provide a concrete implementation of the MDD module. It comprises a denoiser network with a U-net \cite{unet} structure and a diffusion process. Some important parameter settings are shown in Table \ref{info_mdd}.

\begin{table}[!t]
\centering
\resizebox{\columnwidth}{!}{%
\begin{tabular}{c|c|c}
\bottomrule      \textbf{Simulation} &                           \textbf{Parameter}  &             \textbf{Value}   \\ \hline
\multirow{7}{*}{Fog Simulation \cite{fogsim}} &gamma &0.000001\\
&alpha&0.06\\
&noise\_variant&v2\\
&noise&10\\
&r\_noise&random(1, 20)\\
&max\_intensity&255\\ \hline\hline
\multirow{8}{*}{Snow  Simulation \cite{snowsim}} &num\_intervals &64\\
&interval\_index &random(1,64)\\
&snowfall\_rate&0.5\\
&terminal\_velocity&0.2\\
&noise\_floor&0.7\\
&beam\_divergence&0.003\\
&max\_intensity&255\\
\toprule

\end{tabular}%
}
\caption{The detailed configuration of weather simulation. The parameter names refer to the naming of the official source code and the exact meanings can be found in \cite{fogsim,snowsim}.}
\label{info_sim}
\end{table}

\subsection{Configuration of weather simulation}
To help the reader gain a deeper understanding of the severe weather portion of the study in our V2X-R work. As shown in Table \ref{info_sim}, we list here some important configurations for fog and snow simulations, mainly parameters used to adjust the level of adverse weather. Most of the other configurations implemented refer to the default configurations available on their official open-source code \footnote{\href{https://github.com/MartinHahner/LiDAR_fog_sim}{Fog Simulation Code}} \footnote{\href{https://github.com/SysCV/LiDAR_snow_sim}{Snow Simulation Code}}. In addition, the simulation code we implemented will be included in the publicly released code in the future.

\subsection{K-Radar dataset and evaluation metrics}
The K-Radar dataset \citep{K-radar} contains 58 sequences with 34944 frames of 64-line LiDAR, camera, and 4D radar data in various weather conditions. According to the official K-Radar split, we used 17458 frames for training and  17536 frames for testing. We adopt two evaluation metrics for 3D object detection: $AP_{3D}$ and $AP_{BEV}$ of the class "Sedan" and "Bus" at IoU = 0.3. We use the newest version (v2.1) of the label.

\begin{figure*}[!t]
\centering
  \includegraphics[width=0.8\textwidth]{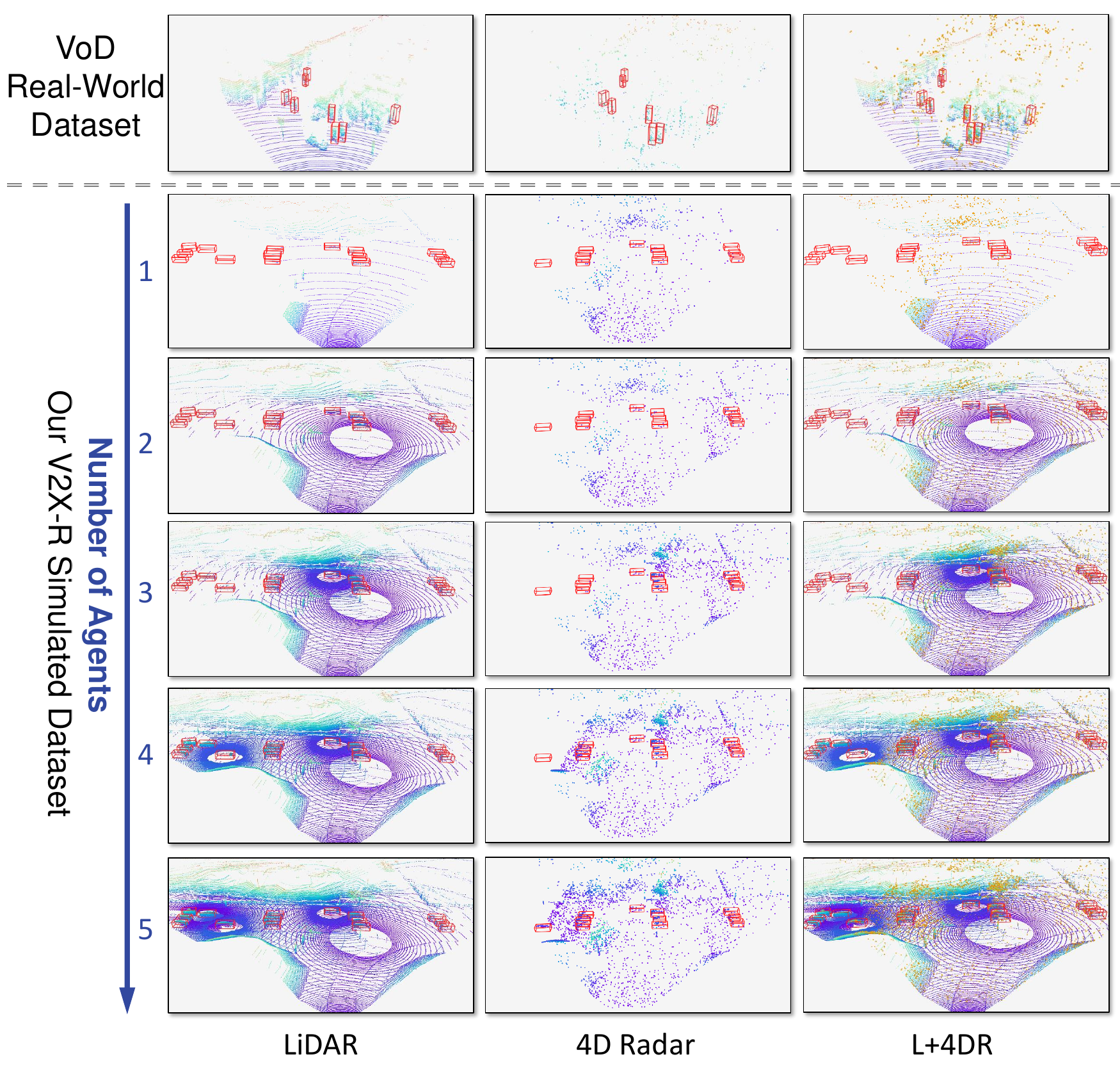}
  
  \caption{Visualization of our V2X-R dataset and VoD \cite{vod} real-world dataset. The L+4DR in the last column indicates that the LiDAR point cloud is visualized together with the 4D radar point cloud, where to distinguish between them, we use colored dots (slightly smaller) for the LiDAR point cloud and orange dots (slightly larger) for the 4D radar point cloud. Colored point clouds are assigned by z-axis values.}
  \label{radar}
\end{figure*}

\section{Visualization of V2X-R Dataset}
Finally, in order to visually intuitively verify the realism of the simulated LiDAR-4D radar data on our V2X-R dataset and the advantages of the cooperative LiDAR-4D radar point cloud. We have visualized and compared the simulated LiDAR-4D radar point cloud on our V2X-R dataset with the real LiDAR-4D radar point cloud on the VoD dataset~\cite{vod}. As shown in Fig. \ref{radar}, it can be found that our simulated LiDAR-4D radar point cloud has a certain degree of realism. This proves the value of conducting 4D radar-related research on our V2X-R dataset. Meanwhile, by comparing the real single-agent 4D radar with the multi-agent 4D radar, it can be found that the multi-agent collaborative 4D radar has a significantly higher resolution. As we introduced in the Introduction section of the main text, the multi-agent cooperative 4D radar has a certain independent perception ability.